\documentclass[conference]{IEEEtran}
\IEEEoverridecommandlockouts

\def\BibTeX{{\rm B\kern-.05em{\sc i\kern-.025em b}\kern-.08em
    T\kern-.1667em\lower.7ex\hbox{E}\kern-.125emX}}

\usepackage{cuted}
\usepackage{times}
\usepackage{url}
\usepackage{array}
\usepackage{textcomp}
\usepackage{stfloats}
\usepackage{verbatim}
\usepackage{setspace}
\usepackage{microtype}
\usepackage{bm}
\usepackage{lipsum,multicol}
\usepackage[mathscr]{euscript}
\usepackage{eepic, amssymb,latexsym, setspace}
\usepackage{rotating}
\usepackage{nicefrac}
\usepackage[parfill]{parskip}
\usepackage{caption}
\usepackage{subcaption}
\usepackage{float}
\usepackage{times}
\usepackage[shortcuts]{extdash}
\usepackage[utf8]{inputenc} 
\usepackage[T1]{fontenc}    
\usepackage{url}            
\usepackage{amsfonts}       
\usepackage{nicefrac}       
\usepackage{microtype}      
\usepackage{xcolor}         
\usepackage{cite}

\usepackage{amsmath,amssymb,amsfonts}
\usepackage{algorithmicx}
\usepackage[linesnumberedhidden,ruled,vlined]{algorithm2e}   

\usepackage{graphicx}
\usepackage{textcomp}
\usepackage{rotating}
\usepackage{multirow}
\usepackage[english]{babel}
\usepackage{float}
\usepackage{mathtools,eqparbox}
\usepackage{multirow}
\usepackage{tikz}

\newcommand{\cN}{\mathcal{N}}

\newcommand{\vvec}{\operatorname{vec}}

\newcommand{\x}{\mathbf{x}}
\newcommand{\y}{\mathbf{y}}

\newcommand{\e}{\mathbf{e}}

\def\bfa{{\mathbf a}}

\def\bfe{{\mathbf e}}

\def\bfx{{\mathbf x}}

\def\bfB{{\mathbf B}}

\def\bfE{{\mathbf E}}

\def\bfH{{\mathbf H}}

\def\bfQ{{\mathbf Q}}
\def\bfR{{\mathbf R}}

\def\bfU{{\mathbf U}}

\newcommand{\mbR}{\mathbb{R}}

\begin{document}

\title{Repair Brain Damage: Real-Numbered Error Correction Code for Neural Network
\thanks{The work of Weiyu Xu was supported by NSF under grants ECCS-2133205.}
}

\author{\IEEEauthorblockN{Ziqing Lu}
\IEEEauthorblockA{\textit{Dept. of AMCS} \\
\textit{University of Iowa}\\
Iowa City, IA, USA \\
ziqing-lu@uiowa.edu}
\and
\IEEEauthorblockN{Myung Cho}
\IEEEauthorblockA{\textit{Dept. of ECE} \\
\textit{California State University}\\
Northridge, CA, USA \\
michael.cho@csun.edu}
\and
\IEEEauthorblockN{Qiutong Jin}
\IEEEauthorblockA{\textit{Dept. of EECS} \\
\textit{University of California}\\
Berkeley, CA, USA \\
qiutong-jin@berkeley.edu}
\and
\IEEEauthorblockN{Weiyu Xu}
\IEEEauthorblockA{\textit{Dept. of ECE} \\
\textit{University of Iowa}\\
Iowa City, IA, USA \\
weiyu-xu@uiowa.edu}
}

\maketitle

\begin{abstract}
We study neural networks (NNs) operating under hardware-induced memory faults and computational errors, which can degrade reliability in safety-critical applications. To address this challenge, we propose a novel real-valued error correction coding (ECC) framework capable of detecting and correcting both types of errors. Our approach imposes structured real-valued linear constraints on network weights, enabling efficient error detection and recovery without increasing the number of parameters or compromising classification performance. The proposed method leverages linear programming and sparse recovery techniques to achieve low-complexity error correction. To the best of our knowledge, this is the first real-number–based ECC scheme that simultaneously addresses memory and computational errors in neural networks.
\end{abstract}


\section{Introduction}
\label{sec:intro}
Deep neural networks have become increasingly important for a wide range of information processing tasks, including classification, regression, image reconstruction, and generative AI. They are also widely deployed in safety-critical applications such as self-driving cars, drones, and robotics \cite{do2018real,ni2020survey,el2024real}. In these settings, ensuring robustness to input errors is essential for safe operation. To this end, a substantial body of research has focused on improving the robustness of neural networks against input perturbations, particularly adversarial examples \cite{bahramali2021robust,wang2019assessing,jakubovitz2018improving,liu2022generating}. Such robustness ensures reliable performance under diverse or perturbed inputs, provided that the neural network itself operates correctly.

In this paper, we consider a different notion of robustness: resilience to a neural network's own failures. In practice, neural networks may suffer from both memory and computational errors. Model parameters, such as weights and biases, are typically stored in non-volatile memory (e.g., flash memory) and may become corrupted over time due to hardware faults, including data retention and memory cell endurance issues in flash memory \cite{cai2015data}. In addition, errors can arise during layer-wise computations (e.g., under low-power or electromigration conditions) due to faulty hardware units, communication noise, or distributed processing. Such errors are particularly prevalent in distributed systems or non-electronic computing platforms \cite{chen2023overview,sui2020review,kalthoff2025acoustic}, which are more susceptible to noise and interference. Moreover, during large-scale training of models such as large language models (LLMs), it is often necessary to detect and correct intermediate computational errors in a timely manner \cite{jiang2025l4,wang2023gemini,Rich2021traning,he2023understanding}.

To address errors in neural networks, numerous studies have explored the use of error correction codes (ECCs) in deep neural networks (DNNs) to improve robustness, security, and reliability. In general, errors in DNNs can be broadly categorized into two types: datapath errors (i.e., computational faults) and memory errors (i.e., storage faults). Existing works have largely focused on applying ECC techniques to mitigate these two types of errors separately. 

Prior works \cite{Huang2020Functional,Ahmed2024NN,Burel2021Zero,Ruiz2024Zero,guan2019place,lee2022value} have primarily focused on mitigating weight parameter errors caused by memory faults. For example,
\cite{Huang2020Functional} proposed using ECCs to protect DNN weights, leveraging deep reinforcement learning to balance redundancy and performance. Similarly, \cite{Ahmed2024NN} embedded ECC parity bits directly into weight parameters during training, enabling correction of memory-induced errors at inference. In addition, \cite{qin2017robustness} introduced a weight nulling scheme that mitigates memory errors by setting detected faulty weights to zero. Another line of works \cite{Burel2021Zero,Ruiz2024Zero,guan2019place,lee2022value} exploits unused or less significant bits in weight representations to store ECC redundancy. For instance, \cite{Burel2021Zero} enforces parity by modifying the least significant bit of each weight, while \cite{guan2019place} repurposes unused bits for redundancy. Overall, these methods operate at the level of individual weights or small groups of weights, aiming to detect and/or correct local errors caused by memory faults. Conceptually, they adapt conventional ECC techniques from memory systems by embedding redundancy alongside weight parameters. However, these approaches primarily address memory faults and do not account for errors arising during computation or communication (i.e., datapath errors).

On the other hand, several studies have addressed datapath errors. In particular, \cite{yu2024error,Deng2010Applying,Verma2019Error,Yu2023COLA,Gupta2022Scalable,dietterich1994solving} explored Error-Correcting Output Codes (ECOCs), which apply ECC principles at the final layer of a DNN. In this approach, the network output is represented as a binary code for each class, enabling correction of errors propagated through the datapath. However, since error correction is applied only at the output layer, ECOCs offer limited protection against memory-induced weight errors, which may accumulate and propagate throughout the network. Other works have investigated fault behavior and mitigation strategies for datapath errors. For example, \cite{Li2017Understanding} studied fault injection in DNN accelerators to analyze the impact of datapath faults, while \cite{Zhang2018Analyzing} examined faults in multiplication-and-accumulation (MAC) operations and proposed a fault-aware pruning strategy to improve resilience. In addition, \cite{raviv2020codnn} introduced a coding-based approach that embeds binary ECCs into inputs or intermediate layers, providing provable robustness to noise and faults without retraining. Since both memory and datapath errors can significantly degrade DNN performance, it is essential to address them jointly.

In this paper, we propose a real-number–based error correction coding (ECC) framework to jointly enhance neural network robustness against both memory and computational errors. The key idea is to impose structured constraints on network parameters and intermediate outputs, introducing redundancy without increasing the number of model parameters. This enables efficient, low-complexity error detection and supports simultaneous correction of both error types. Compared with prior work \cite{Huang2020Functional,Ahmed2024NN,Burel2021Zero,Ruiz2024Zero,guan2019place,lee2022value}, the proposed approach allows the number of correctable errors to scale linearly with the number of parameters \cite{candes2005decoding}, whereas existing methods (e.g., \cite{Ahmed2024NN}) are limited to correcting a constant number of errors due to short code lengths. Moreover, unlike the previous approaches, our method does not require additional weight parameters or neurons, as it imposes structured constraints directly on the trained weights and biases. To the best of our knowledge, this is the first real-valued ECC framework that can jointly detect and correct both memory and computational errors in neural networks.


The remainder of this paper is organized as follows. Section \ref{sec:prob} introduces notation and terminology, followed by problem formulation and our proposed approach. Section \ref{sec:propose} presents the algorithm for detecting and correcting memory and computational errors. Section \ref{sec:simulation} provides numerical results to validate the proposed method. Section \ref{sec:conclusion} concludes the paper.

\textbf{Notations and network terminologies:} 
Let a neural network based classifier $G(\cdot):\mathbb{R}^d\to\mathbb{R}^k$ be implemented through a $l$-layer NN which has $l-1$ hidden layers and has $l+1$ columns of neurons (including the neurons at the input layer and output layer). We denote the number of neurons at the inputs of layers $1$, $2$, ..., and $l$ as $n_1$, $n_2$, ...., and $n_l$ respectively.  At the output of the output layer,  the number of neurons is $n_{l+1}=k$, where $k$ is the number of classes. We define the bias terms in each layer as $\boldsymbol{\delta_{1}}\in\mbR^{n_{2}}, \boldsymbol{\delta_{2}}\in\mbR^{n_{3}}, \cdots, \boldsymbol{\delta_{l-1}}\in\mbR^{n_{l}}, \boldsymbol{\delta_{l}}\in\mbR^{n_{l+1}}$, and the weight matrices $\bfH_{i}$ for the $i$-th layer is of dimension $\mathbb{R}^{n_{i+1} \times n_i}$. The element-wise activation functions in each layer are denoted by $\sigma(\cdot)$, and some commonly used activation functions include ReLU and leaky ReLU \cite{szandala2020review}. Hence, when the input is $\x$, the output $\y$ of a neural network is given by $\y=G(\x) 
= \sigma(\bfH_{l}\sigma(\bfH_{l-1} \cdots \sigma(\bfH_{1}\x + \boldsymbol{\delta_{1}})  \cdots + \boldsymbol{\delta_{l-1}})+ \boldsymbol{\delta_{l}}).$


\section{Problem Formulation}
\label{sec:prob}

We present our ECC framework using a feedforward NN classifier, although the proposed approach can be extended to other network architectures. Our ECC framework is applicable to the full set of weights and biases across the network. However, for simplicity of exposition, we focus on the ECC design for the weights of a single layer.

Suppose that one NN layer's weights are $\bfH \in \mbR^{p \times q}$, and that the input to this layer is $\x \in \mbR^{q \times 1}$. If there are no memory or computational errors, the output of this layer will be $\y=\bfH \x$. Sometimes, the weight $\bfH$ can be contaminated with memory error $\bfE \in \mbR^{p \times q}$ and thus the actual stored weights are $\tilde{\bfH}=\bfH+\bfE$.  Moreover, there can be computational errors $\e  \in \mbR^{p}$. The output of this layer under the memory and computational errors is\\[-20pt]
\par\noindent\small
\begin{align*}
    \y=\tilde{\bfH} \x+\e=(\bfH+\bfE)\x+\e.
\end{align*}
\normalsize
The goal of this work is to develop efficient methods for detecting computational and/or memory errors during neural network inference with minimal overhead in terms of time and computational cost, while being sufficiently powerful to correct both types of errors across the entire network. A key target application is AI-driven robotic systems operating in harsh environments (e.g., Mars rovers or robots in radioactive settings), where reliable and uninterrupted operation is critical despite adverse conditions.





\section{Linear Programming-based Error Correction Code for Error Tolerant DNN}
\label{sec:propose}
Our goals for designing ECC in DNNs are to achieve strong error-correction capability, low-complexity error detection, minimal (or no) additional memory overhead for ECC storage, and the ability to handle both memory and datapath errors simultaneously. To meet these objectives, we propose a linear programming-based error detection and correction scheme. In the following subsections, we describe how to construct a DNN that supports this scheme, and  present the detection and correction processes in detail.

\subsection{Constraints for error detection/correction capability}
For ECC, we introduce two types of linear constraints for the weights: the 1st type of constraints are general linear constraints over  elements of $\bfH$, and the 2nd type of constraints are a special type of linear constraints over rows of $\bfH$. The 2nd type of constraints are specifically constructed to detect and correct the computational errors, while the 1st and 2nd types of constraints can be jointly used to detect and correct memory errors. 

For the 1st type of linear constraints, we introduce $t$ matrices  $\bfB_j \in \mbR^{p \times q}$, $1\leq j \leq t$. The code requires, for every $j$,  \\[-20pt]
\par\noindent\small
\begin{align}
\label{eq:condition2}
\langle \bfB_j, \bfH \rangle =0.
\end{align}
\normalsize
Namely, the inner product between $\bfB_j$ and $\bfH$ is 0. These $t$ constraints are thus general linear constraints for $\bfH$.  

For the 2nd type of constraints, we introduce $s$ vectors $\bfa_i \in \mbR^{p}$, $1\leq i \leq s$.  We would require that for each $\bfa_i$, we have $\bfa_i^T \times \bfH= 0_{1\times q}$. Thus, when there are no errors in the observations or memories, for each  $\bfa_i$,
\par\noindent\small
\begin{align}
 \bfa_i^T \y=\bfa_i^T \bfH\x=(\bfa_i^T \bfH) \x=0.
\label{eq:errordetectioncondiiton}
\end{align}
\normalsize
The 2nd type of constraints impose linear structures on the output $\y$. From another perspective, these 2nd type of constraints also impose $s \times q$ linear constraints on the weights, similar to the 1st type of constraints. However, these 2nd type of linear constraints are special enough to guarantee there are linear structures in the output $\y$, while the 1st type of constraints cannot guarantee those structures. 

\textbf{Remarks}:  To accommodate the biases $\boldsymbol{\delta}$ at that layer's output neurons, we can incorporate $\boldsymbol{\delta}$ with $\bfH$, and obtain the input to the activation functions as:  \\[-20pt]
\par\noindent\small
\begin{align*}
\y= \begin{pmatrix} \bfH, \boldsymbol{\delta} \end{pmatrix} \begin{pmatrix} \x \\
                                                                  1 \end{pmatrix}.
\end{align*}
\normalsize
 We can then treat $\begin{pmatrix} \bfH, \boldsymbol{\delta} \end{pmatrix}$ as a new matrix $\bfH_{new}$ of dimension $p \times (q+1)$, and construct 1st and 2nd types of constraints for $\bfH_{new}$ similarly.

\subsection{LP-based error detection and correction}
The low-complexity error detection can be performed using the condition in \eqref{eq:errordetectioncondiiton}. Specifically, by checking whether the product of the layer output $\y$ and the constraint vectors $\bfa_i$ yields zero, one can efficiently detect the presence of errors, regardless of their type (memory fault or datapath fault). This is because under very general conditions, the errors which satisfy these constraints can be shown to have measure $0$.

If errors are detected, the algorithm proceeds to error correction. Assuming that memory errors are sparse, we formulate the following $\ell_1$ minimization problem to identify the memory error $\bfE$:\\[-20pt]
\par\noindent\small
\begin{align}
\label{eq:L1_minimization}
    &\underset{\bfE \in \mbR^{p \times q} }{\text{minimize}}\;\| \bfE \|_1 \nonumber\\
    &\text{subject to}\;\;\;\bfa_i^T (\tilde{\bfH} - \bfE) = 0_{1 \times q}, \;\; \forall i, \nonumber\\
    &\quad\quad\quad\quad\;\; \langle \bfB_j, \tilde{\bfH} - \bfE \rangle = 0, \;\; \forall j,
\end{align}
\normalsize
where $\tilde{\bfH}$ is the weight matrix corrupted by memory errors, i.e., $\tilde{\bfH} = \bfH + \bfE$, $\|\bfE\|_1$ is the $\ell_1$ norm of $\bfE$ (by treating it as a matrix-shaped vector). Once $\bfE$ is identified by solving the above optimization problem using LP, the clean weight matrix can be recovered as $\bfH = \tilde{\bfH}-\bfE$. The layer output can then be corrected in the following manner. 

Let ${\y}$ denote the corrupted layer output due to memory error $\bfE$ and/or datapath error $\e$, i.e., ${\y} = (\bfH + \bfE) \x + \e = \tilde{\bfH} \x + \e$. The correct output is then given by $\y_{c} = \bfH \x = {\y} - \bfE \x - \e$, where $\bfE$ is obtained from solving the minimization problem, $\e$ can be directly computed as ${\y}- \tilde{\bfH} \x$, and both ${\y}$ and $\x$ are known. Algorithm \ref{alg:LP_ECC} summarizes the proposed linear programming-based error detection and correction method for a single layer. For multi-layer settings, the algorithm can be applied to each layer independently.

\begin{algorithm}[t]
  \caption{Linear Programming-Based Error Detection and Correction Algorithm in a Layer}
  \label{alg:LP_ECC}
  \SetAlgoLined
{\small
   \KwIn{ $\y$, $\tilde{\bfH}$, $\{\bfa_i\}_{i=1}^s$, and $\{\bfB_i\}_{i=1}^t$}
   \tcc{Error Detection \& Correction Process}
   \uIf { $\bfa_i^T \y = 0$, $\forall$ i, } 
   {
        Declare no error \par
    }
    \Else
    {
   			\tcc{Error Correction Process}
            Solve \eqref{eq:L1_minimization} \par
            Correct the weight matrix (i.e., $\bfH = \tilde{\bfH} - \bfE$) \par
            Calculate $\e = \y - \tilde{\bfH} \x$ \par
            Correct the layer output (i.e., $\y_{c} = {\y} - \bfE \x - \e$)
    }%
  \KwOut{Corrected weight matrix $\bfH$ and layer output $\y_{c}$ }
}%
\end{algorithm}
Alternatively, one can have the following more sophisticated  optimization problem to jointly find out $\bfH$ and $\e$ directly:
\par\noindent\small
\begin{align}
\label{eq:combined_L1_minimization}
    &\underset{\bfE \in \mbR^{p \times q}, \e \in \mbR^{p \times 1}}{\text{minimize}}\;\| \bfE \|_1 +\|\e\|_1\nonumber\\
    &\text{subject to}\;\;\;\bfa_i^T (\tilde{\bfH} - \bfE) = 0_{1 \times q}, \;\; \forall i, \nonumber\\
    &\quad\quad\quad\quad\;\; \langle \bfB_j, \tilde{\bfH} - \bfE \rangle = 0, \;\; \forall j, \nonumber\\
    &\quad\quad\quad\quad\;\;\y=\tilde{\bfH}\x+\e.
\end{align}
\normalsize

In the following subsection, we describe how to construct a DNN (i.e., its weight matrices) to enable the linear programming-based error detection and correction. 

\subsection{Constructing a DNN to enable the linear programming-based error detection and correction}
\label{subsec:ecc_projection}
The linear programming-based error detection and correction ability is obtained through conditions \eqref{eq:errordetectioncondiiton} and \eqref{eq:condition2}. To achieve these conditions, for a weight matrix $\bfH$ (which we can obtain through training), we reconfigure it to impose conditions \eqref{eq:errordetectioncondiiton} and \eqref{eq:condition2} either during or after training. 

To have the conditions in a weight matrix, we can vectorize a weight matrix of size $p \times q$ into a vector of size $pq \times 1$. We can impose linear relationships among the associated weights.  As explained earlier, Condition \eqref{eq:errordetectioncondiiton} is a special form of Condition \eqref{eq:condition2}. So we just discuss how to obtain weights satisfying Condition \eqref{eq:condition2}. 

For Condition \eqref{eq:condition2}, with a vectorized weight matrix sized in $pq \times 1$, we can generate $k$ (deterministically or randomly generated) constraint vectors, denoted by $\bfU \in \mbR^{pq \times k}$, and orthonormalize it by using QR decomposition to form a basis for the constraint subspace. Namely, $\bfU=\bfQ \bfR$. The original weight vector (i.e., vectorized weight matrix) is then projected onto the orthogonal complement of this subspace as\\[-20pt]
\par\noindent\small
\begin{align}
\label{eq:projection_ecc}
    \vvec(\bfH) = \vvec(\bfH_o) - \bfQ(\bfQ^T)\vvec(\bfH_o),
\end{align}
\normalsize
ensuring that the resulting weights $\bfH$ satisfy all imposed linear constraints, (i.e., the condition \eqref{eq:condition2}). Here, $\bfH_o$ represents a weight matrix not having the condition \eqref{eq:condition2}, and $\vvec(\cdot)$ vectorizes a matrix by column-stacking. This projection preserves the dimensionality of the network and does not introduce additional trainable parameters. 

Finally, the projected weights are reshaped back to their original matrix form, written back into the network, and the updated model parameters are saved. This procedure imposes real-number-based structures into the network weights, enabling robustness against memory and computational errors while maintaining the original network architecture and performance. 

\textbf{Remarks:} The structures imposed on the NN weights can be enforced during training or after training.



\subsection{Advantages of the proposed method}
Conventional error correction codes (ECCs), such as LDPC and BCH codes \cite{lin2004error}, are primarily designed for binary representations. In contrast, the proposed framework operates directly in the real domain, which is more natural for neural network parameters and computations. As a result, the error correction procedure relies on solving an $\ell_1$ minimization problem, which can be computationally more demanding than conventional binary ECC decoding. However, this increased computational cost enables greater flexibility and stronger recovery guarantees, particularly in handling sparse memory and computational errors in high-dimensional real-valued parameters.

Despite this trade-off, the proposed method offers several important advantages. First, unlike conventional ECC schemes that require dedicated redundancy bits, the proposed approach introduces structured constraints without increasing the number of model parameters. This preserves the original network architecture. Second, the constraint matrices $\mathbf{B}_j$'s can be shared across multiple weight matrices and layers. In other words, the same set of constraints can be reused to recover different weight matrices $\mathbf{H}$'s throughout the network. This reusability can reduce the memory required to store redundancy information, making the approach more scalable for large models.

Furthermore, the proposed framework supports efficient error detection through simple linear checks, allowing errors to be identified with low computational overhead before invoking more expensive correction procedures. In addition, the method can be applied post hoc to trained models, eliminating the need for retraining or architectural modifications. These properties make the proposed approach particularly attractive for deployment in resource-constrained or unreliable hardware environments, where both memory and computational faults may occur.

\section{Numerical Experiments}
\label{sec:simulation}
In order to assess the performance of our proposed method, we performed experiments  with the MNIST dataset and the CIFAR-10 dataset.  

\subsection{MNIST Dataset}
In this numerical experiment, we consider a feedforward DNN consisting of three layers: one input layer, one hidden layer, and one output layer. The weight matrices have dimensions $256 \times 784$, $128 \times 256$, and $10 \times 128$ for the input, hidden, and output layers, respectively. The ReLU activation function is used for the input and hidden layers. In the output layer, a softmax function is applied to classify 10 different digits.

As a baseline, we train the feedforward DNN using Stochastic Gradient Descent (SGD) with a learning rate of 0.01. The cross-entropy loss function is employed, and the network is trained for 100 epochs. 

For the projected DNN (referred to as ECC-DNN), we project the weight parameters to incorporate error detection and correction capability using the method described in Section \ref{subsec:ecc_projection}. Although the ECC structure can be applied to all layers, for simplicity, we apply it only to the first (input) layer. To reduce the computational complexity of the optimization problem in \eqref{eq:L1_minimization}, we partition the 784 columns of the weight matrix into 56 groups, each containing 14 columns. For each group, we impose the constraint that the last row is equal to the negative of the sum of the preceding 255 rows. In addition, we further introduce constraints corresponding to \eqref{eq:condition2} through \eqref{eq:projection_ecc}, where $k=500$.

For the noise and errors in the signal model, i.e., $\y=(\bfH + \bfE)\bfx + \bfe$, the memory error matrix $\bfE$ is similarly partitioned into 56 groups by grouping 14 columns per group. Consequently, each submatrix has dimensions $256 \times 14$. For each submatrix, we randomly generate a sparse error matrix, where the locations of the non-zero elements are chosen uniformly at random and the error values are drawn from $\cN(0,2^2)$. The number of errors per group varies from 5 to 30, resulting in a total number of errors in $\bfE$ ranging from 70 to 420. For the datapath (or communication) error vector $\bfe$, we generate a sparse vector with two non-zero elements, whose values are drawn from $\cN(0, 0.01^2)$. 

\begin{figure}[t]
    \centering
    \includegraphics[width=0.35\textwidth]{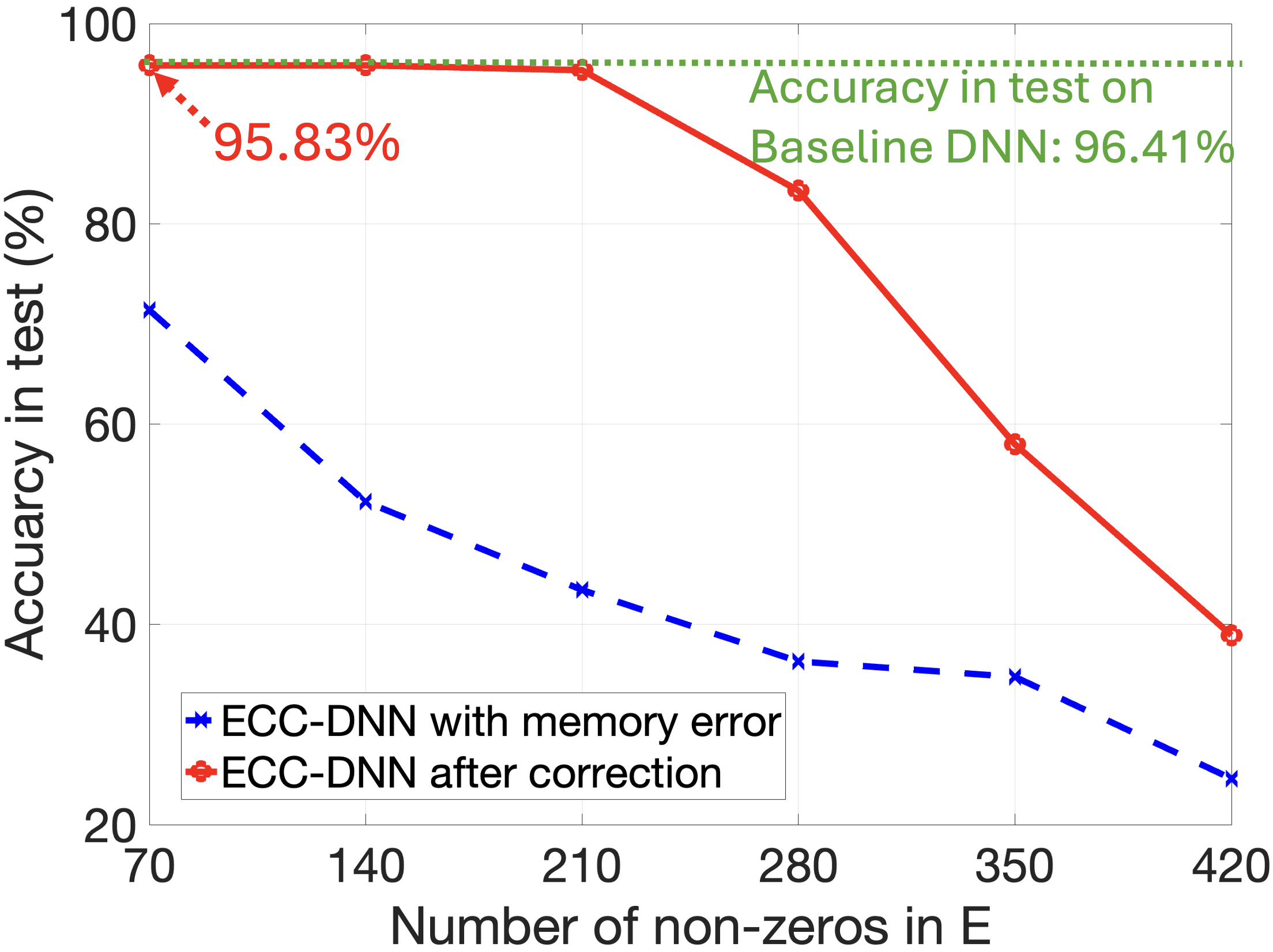}
    \caption{\small{Numerical results of MNIST test classification accuracy for ECC-DNN with and without error correction under varying numbers of weight errors.}}
    \label{fig:mnist_ecc_result}
\end{figure}

To the best of our knowledge, this is the first work to jointly consider both computational and memory errors. We evaluate the performance of ECC-DNN under both error types by comparing results before and after error correction. To obtain statistical results, we conduct 10 independent trials for each noise configuration and report the average test classification accuracy. Figure \ref{fig:mnist_ecc_result} presents the results, where the x-axis denotes the number of errors in $\mathbf{E}$ and the y-axis shows the average test accuracy over 10 trials. The red solid and blue dotted curves correspond to ECC-DNN with and without error correction, respectively. The results show that, for up to 210 errors out of $256 \times 784$ weight parameters, the ECC-DNN maintains nearly the same test accuracy as the baseline (i.e., without errors and without weight projection), with negligible performance degradation. Additionally, the results demonstrate that the proposed method can provide strong error resilience even in the presence of substantial errors.

\subsection{CIFAR-10 Dataset}

We further perform numerical experiments over the CIFAR-10 dataset. We consider the WideResNet-22-8\cite{zagoruyko2016wide} that has 22 convolution layers followed by an average pooling and a fully-connected output layer. Batch normalization, ReLU, and dropout are included in each residual block.

As a baseline, we train WideResNet-22-8 using sharpness-aware minimization (SAM) optimizer with a learning rate of $0.1$. We use the entropy loss function with label smoothing and the network is trained for $100$ epochs.

We construct an ECC for the output layer, which can be modeled by a $10 \times 512$ weighted matrix $\bfH$. For the 1st type of constraints, we randomly generate $500$ constraints $\bfB_j$'s, where elements of $\bfB_j$ are independently sampled from $\mathcal{N}(0,1)$. For the 2nd type, we impose the constraint that the last row is equal to the negative of the sum of the preceding $9$ rows. 

We impose the noise and error in the same way as we do for the MNIST. The sparse memory error $\bfE$ has the same dimension as $\bfH$, and its nonzero elements are distributed from $\mathcal{N}(0,2^2)$. The communication error $\bfe$ is a sparse vector with two non-zero vectors, whose values are distributed from $\mathcal{N}(0,0.01^2)$.

\begin{figure}
    \centering
    \includegraphics[scale=0.30]{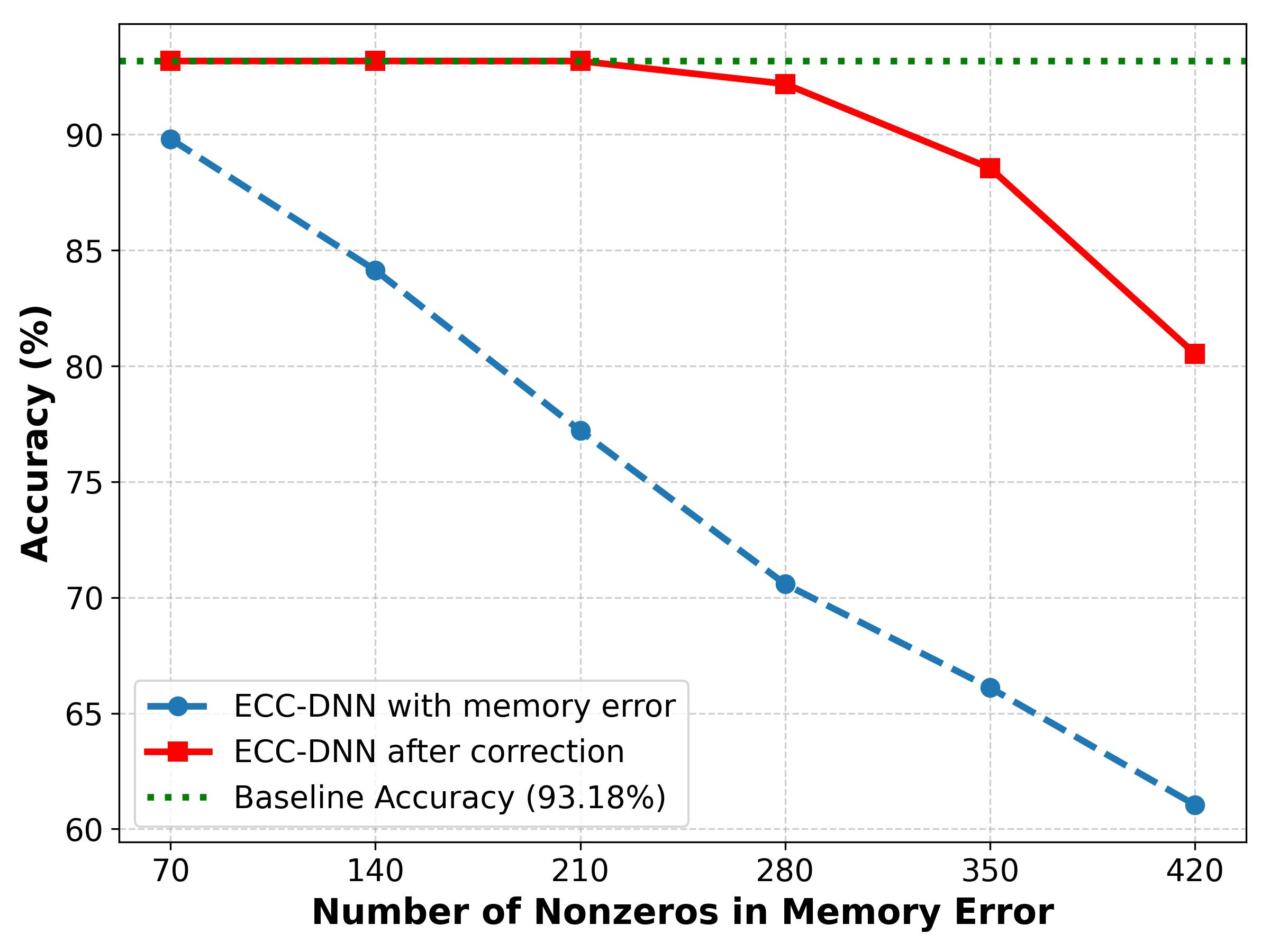}
    \caption{\small{Numerical results of CIFAR-10 test classification accuracy for ECC-DNN, with and without error correction}}
    \label{fig:CIFAR}
\end{figure}

We report the average test classification accuracy of $100$ trials with independent noise configurations. In Figure \ref{fig:CIFAR}, we present the simulation results with and without error correction. The x-axis denotes the number of non-zero elements in $\bfE$ and the y-axis denotes the average test classification accuracy over $100$ independent trials.

The classification accuracy of the baseline model is $93.18\%$. The red solid line and the blue dotted line represent the ECC-DNN with and without error correction, respectively. Up to $210$ errors out of $5120$ parameters, the proposed linear programming-based error correction algorithm maintains a stable test classification accuracy. 
We note that this is the output, each row of the weight matrix means the direction the NN looks at for each class. So imposing the constraint that the 10th row is equal to the negative sum of the preceding 9 rows seems very restrictive or demanding for the NN classifier. It is quite surprising this constraint can provide nice error detection capability while not compromising classification accuracy.

\section{Conclusion}
\label{sec:conclusion}
In this work, we studied neural networks operating in the presence of memory faults and computational errors (i.e., datapath faults). We proposed a novel real-number–based error correction coding framework that enables the detection and correction of both types of errors within a unified formulation. By introducing carefully designed real-valued linear constraints on network weights and intermediate computations, the proposed method provides reliable error detection and correction without increasing the number of model parameters or modifying the network architecture. The proposed framework offers several key advantages. It jointly addresses memory and computational errors, supports low-complexity error detection, and enables the number of correctable errors to scale with the model size. In addition, the framework can be applied post hoc to trained networks, and the reuse of constraint matrices across layers reduces the memory overhead required for redundancy. Numerical results demonstrate that the proposed method maintains classification performance even in the presence of substantial errors. Overall, the proposed approach provides a practical and scalable solution for improving the reliability of neural networks in error-prone hardware environments. Future work includes extending the framework to broader network architectures and improving the efficiency of the optimization-based error correction procedure.


\bibliographystyle{IEEEbib}
\bibliography{refs}

\end{document}